\documentclass{article}

\usepackage{microtype}
\usepackage{graphicx}
\usepackage{subcaption}
\usepackage{booktabs} 
\usepackage{bbm} 
\usepackage{hyperref}
\usepackage{booktabs}
\usepackage{multirow}
\usepackage{colortbl}
\usepackage{xcolor}
\usepackage{enumitem}
\usepackage{makecell}
\usepackage{fontawesome5}
\usepackage{pifont}
\usepackage{adjustbox}   
\usepackage{tcolorbox}   
\usepackage{amssymb}     
\usepackage{pifont}      
\tcbuselibrary{skins,breakable}  

\definecolor{cvprblue}{rgb}{0.21,0.49,0.74}
\definecolor{lightgreen}{HTML}{CCFFCC} 
\definecolor{lightyellow}{HTML}{FFFFCC} 
\definecolor{lightgrey}{HTML}{DDDDDD} %
\newcommand{\eg}{\emph{e.g.,}}

\usepackage[preprint]{icml2026}
\makeatletter
\renewcommand{\printAffiliationsAndNotice}[1]{\global\icml@noticeprintedtrue%
  \stepcounter{@affiliationcounter}%
  {\let\thefootnote\relax\footnotetext{\hspace*{-\footnotesep}\ificmlshowauthors #1\fi%
      \forloop{@affilnum}{1}{\value{@affilnum} < \value{@affiliationcounter}}{
        \textsuperscript{\arabic{@affilnum}}\ifcsname @affilname\the@affilnum\endcsname%
          \csname @affilname\the@affilnum\endcsname%
        \else
          {\bf AUTHORERR: Missing \textbackslash{}icmlaffiliation.}
        \fi
      }.%
      \ifdefined\icmlcorrespondingauthor@text
         { }\icmlcorrespondingauthor@text.
      \fi
    }
  }
}
\makeatother

\usepackage{amsmath}
\usepackage{mathtools}
\usepackage{amsthm}

\usepackage[capitalize,noabbrev]{cleveref}

\theoremstyle{plain}

\theoremstyle{definition}

\theoremstyle{remark}

\usepackage[textsize=tiny]{todonotes}
\usepackage{pifont}
\usepackage{marvosym}

\icmltitlerunning{Thinking with Spatial Code}

\begin{document}

\twocolumn[

\icmltitle{Thinking with Spatial Code for Physical-World Video Reasoning}
  \icmlsetsymbol{equal}{*}

\centerline{\bf Jieneng Chen$^{1,*,\text{\scriptsize \Letter}}$ \;Wenxin Ma$^{1,*}$ \;Ruisheng Yuan$^{1,*}$ \;Yunzhi Zhang$^{2,*}$ 
\;Jiajun Wu$^{2,\dagger}$ \;Alan Yuille$^{1,\dagger}$}
\vspace{0.1in}
\centerline{$^1$Johns Hopkins University \quad $^2$Stanford University  
}
\vspace{0.05in}

  \vskip 0.3in
]
\printAffiliationsAndNotice{$^*$Equal contribution  $^\dagger$Equal advising \quad \text{\Letter}$<$jchen293@jh.edu$>$}

\begin{abstract}
We introduce Thinking with Spatial Code, a framework that transforms RGB video into explicit, temporally coherent 3D representations for physical-world visual question answering. We highlight the empirical finding that our proposed spatial encoder can parse videos into structured spatial code with explicit 3D oriented bounding boxes and semantic labels, enabling large language models  (LLMs) to reason directly over explicit spatial variables.
Specifically, we propose the spatial encoder that encodes image and geometric features by unifying 6D object parsing and tracking backbones with geometric prediction, and we further finetuning LLMs with reinforcement learning using a spatial rubric reward that encourages perspective-aware, geometrically grounded inference. 
As a result, our model outperforms proprietary vision–language models on VSI-Bench, setting a new state-of-the-art. Code is available at \url{https://github.com/Beckschen/spatialcode}.

\end{abstract}    
\section{Introduction}
\label{sec:intro}
\vspace{-2mm}

\begin{figure}[ht!]
  \centering
   \includegraphics[width=1\linewidth]{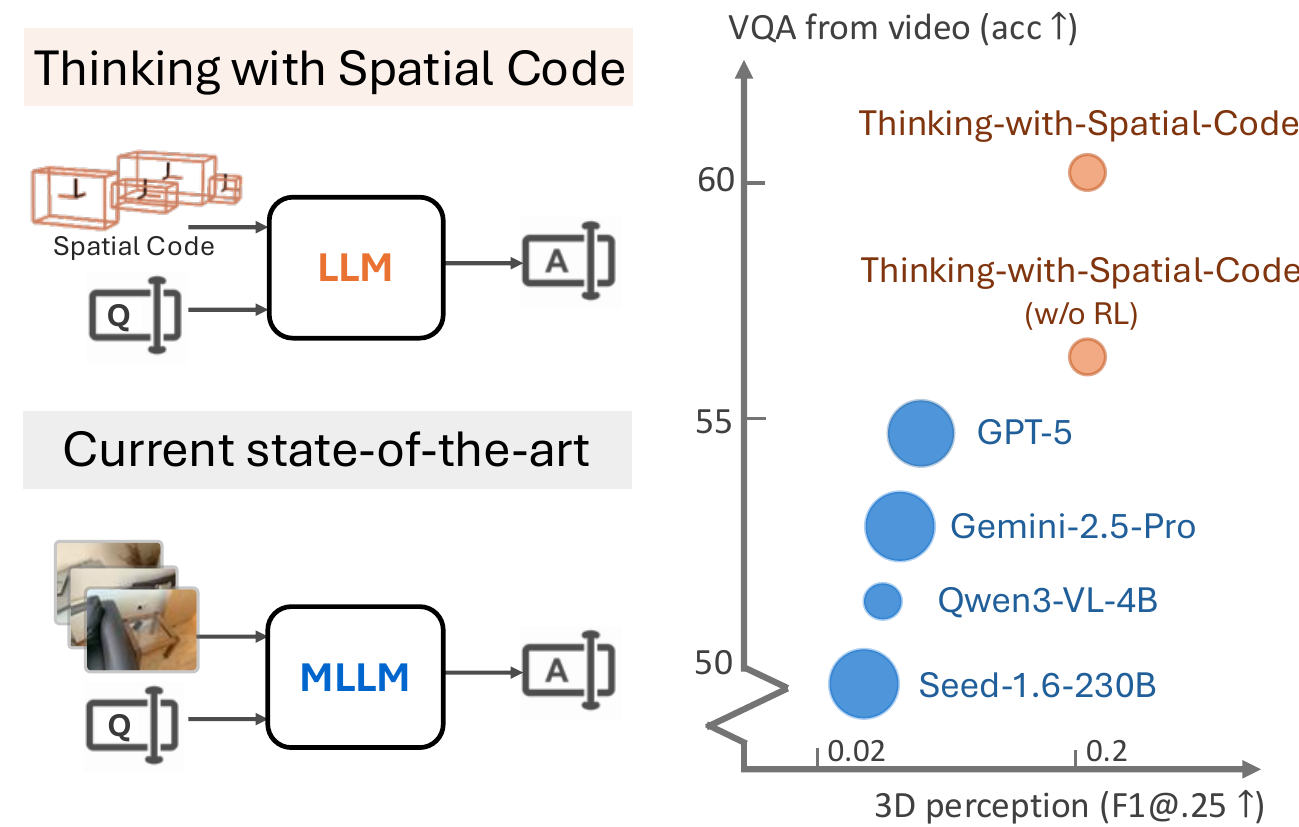}  
   \vspace{-1.3em}
  \caption{
  \textbf{Thinking with Spatial Code enables superior spatial reasoning from video.} \textit{Left:} Unlike current state-of-the-art multimodal LLMs (MLLMs) that reason directly over the raw RGB image or video, our approach first parses video into explicit 3D spatial codes, then prompts a text-only LLM to reason over these symbolic descriptions. \textit{Right:} On VSI-Bench~\citep{yang2025thinking}, our method fine-tuned on Qwen3-4B~\cite{yang2025qwen3} significantly outperforms leading MLLMs including GPT-5o, Gemini-2.5, and Qwen3-VL in video-spatial reasoning accuracy. Reinforcement learning with spatial rubric rewards further improves performance. 
  Dot size indicates model scale (4B--230B parameters; GPT and Gemini sizes are undisclosed).
  This demonstrates that the quality of 3D spatial representation, rather than model scale alone, is the key bottleneck for spatial reasoning.
  }
  \label{fig:teaser}
  \vspace{-5mm}
\end{figure}

Humans continuously perceive the physical world not as a sequence of disjointed frames, but as a coherent 3D environment that unfolds over time. From streams of visual inputs, we effortlessly parse spatial layouts, track objects, infer their dynamics, and reason about causal interactions. Achieving such spatially grounded understanding from video remains a long-standing challenge for machine intelligence. Despite the impressive progress of recent large multimodal models (MLMMs), their reasoning is primarily linguistic and appearance-based, lacking explicit 3D structure or spatial continuity. As a result, they can describe \textit{what} they see but struggle to reason about \textit{where} things are, \textit{how} they are oriented relative to one another, and \textit{when} they disappear and reappear -- abilities essential for physical-world perception.

\begin{figure*}[ht!]
  \centering
     \vspace{-3mm}
   \includegraphics[width=1\linewidth]{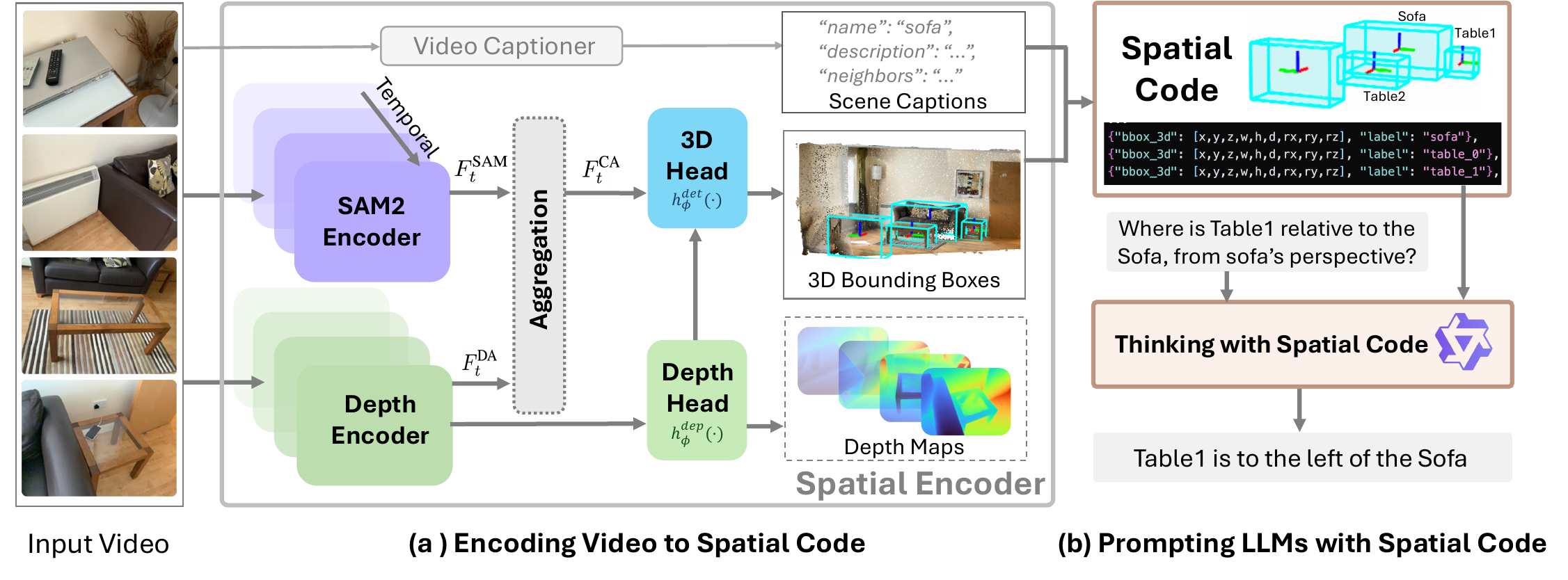}  
   \vspace{-6mm}
  \caption{\textbf{Overview.} \textbf{(a) Encoding Video to Spatial Code:} The Spatial Encoder processes video through a dual-encoder architecture. The SAM-2~\cite{ravi2024sam} encoder extracts object-level features $F_{\text{sam}}$ with temporal attention, while the Depth Encoder (from Depth Anything 3~\cite{lin2025depth}) extracts spatial features $F_{\text{dep}}$. Cross-attention fuses these representations into $F_{\text{ca}}$, which feeds into a 3D Head for predicting 3D object bounding boxes with 3D orientation and a Depth Head for dense geometric supervision. Outputs are structured into symbolic spatial codes encoding object categories, positions, sizes, and orientations. \textbf{(b) Prompting LLMs with Spatial Code:} The spatial codes serve as explicit, interpretable inputs to LLMs for spatial reasoning. Given a query requiring perspective-aware understanding (\eg~``Where is Table1 relative to the Sofa, from sofa's perspective?''), the LLM reasons directly over the structured 3D representations to produce geometrically grounded answers. 
  }
  \label{fig:framework}
   \vspace{-3mm}
\end{figure*}

To bridge this gap, we introduce \textbf{Thinking-with-Spatial-Code}, a framework that transforms RGB video into explicit, temporally coherent 3D representations for physical-world visual question answering. Our key insight is that the quality of spatial representation, rather than model scale alone, is the critical bottleneck for spatial reasoning. We highlight the empirical finding that the spatial encoders trained to parse videos into structured spatial codes — with explicit 3D bounding boxes and semantic labels — perform reliably on real-world distributions, enabling large language models to reason directly over explicit spatial variables.


Our framework consists of two main components. First, a \textbf{Spatial Encoder} transforms streaming video into structured spatial codes, each encoding an object's semantic label, 3D position, size, and orientation. It integrates a dual-encoder architecture combining SAM-2~\citep{ravi2024sam} for object-level features and Depth Anything 3~\citep{lin2025depth} for geometric features, jointly performing segmentation, tracking, and 3D reconstruction. Second, we prompt LLMs with these symbolic spatial codes for spatial reasoning. This design enables LLMs to perform \textit{thinking with spatial code} through explicit coordinate reasoning, while leveraging commonsense knowledge (\eg~understanding that ``the front of a sofa'' implies a canonical facing direction).

To further enhance reasoning capabilities, we employ reinforcement learning
with a novel spatial rubric reward, motivated by careful empirical examination of pre-trained models' behaviors. We observe that models often exhibit a reasoning-action disconnect: correctly analyzing spatial relationships in chain-of-thought but producing incorrect final answers. Our spatial rubric reward evaluates reasoning quality along multiple interpretable dimensions, including perspective-based reasoning, orientation awareness, and directional consistency, while penalizing common failure modes such as viewer-centric errors.

Comprehensive experiments demonstrate that Thinking-with-Spatial-Code achieves state-of-the-art results on VSI-Bench~\citep{yang2025thinking}, outperforming both proprietary MLLMs such as GPT-5o and Gemini-2.5, and open-source models including Qwen3-VL. 

We summarize our contributions as follows:
\begin{itemize}[leftmargin=*,itemsep=2pt,topsep=0pt]
    \item We introduce \textbf{Thinking-with-Spatial-Code}, a new paradigm that parses streaming video into explicit 3D spatial codes for LLM-based inference.
    \item We provide an empirical recipe for training a perception module that unifies dual visual encoding, 6D object parsing and tracking, and geometric densification to generate structured spatial codes from RGB video.
    \item We finetune LLMs for video VQA using spatial code with a novel \textbf{spatial rubric reward} that encourages perspective-aware, geometrically grounded reasoning.
    \item The model achieves state-of-the-art performance on VSI-Bench, 
    and demonstrate the key finding that perception quality is a critical bottleneck for the spatial reasoning performance of MLLMs.
\end{itemize}

We will make our code, models and training recipe fully public to facilitate further research in this direction.
\section{\fullname}
\label{sec:method}

We study physical-world visual question answering (VQA) from videos. Given a spatial query $\mathbf{x}^q$ in the form of text and a RGB video $\mathcal{\mathbf{x}^\text{video}}\in \mathbb{R}^{3\times H\times W\times T}$ of $T$ frames and spatial resolution $H\times W$, the goal is to infer a text responses $\mathbf{y}$. 

Canonical MLLMs approximate the conditional distribution $p(\mathbf{y} \mid \mathbf{x}^\text{video}, \mathbf{x}^q)$, heavily relying on final ground truth answers to learn during training. The sparse nature of such training signals makes the model susceptible to shortcut solutions that rely on 2D appearance cues or viewer-centric biases rather than recovering metric 3D structure. 

We propose to explicitly introduce \textit{spatial codes} $\mathbf{c}$, an explicit intermediate representation that captures 3D structures of scenes.
Specifically, 
\begin{equation}
p(\mathbf{y} \mid \mathbf{x}^\text{video}, \mathbf{x}^{q}) = \int p(\mathbf{y} \mid \mathbf{c}, \mathbf{x}^\text{video}, \mathbf{x}^{q}) \, p(\mathbf{c} \mid \mathbf{x}^\text{video}, \mathbf{x}^q) \, \mathrm{d}\mathbf{c}.
\label{eq:factorization}
\end{equation}
The factorization allows for imposing dense supervision signals on $\mathbf{c}$ (\S\ref{ssec:encoder}), is immediately compatible with the interface of existing large language models (\S\ref{ssec:prompting}), and further benefits the isolation between perception errors and reasoning errors, which motivates reward designs for reinforcement-learning finetuning (\S\ref{ssec:rl}).

In practice, rather than marginalizing over all possible spatial interpretations, we adopt a \textit{perception-then-reasoning} paradigm that commits to a maximum a posteriori estimate, which reduces \cref{eq:factorization} into the following:
\begin{gather}
\mathbf{c}^* = \arg\max_{\mathbf{c}} \, p(\mathbf{c} \mid \mathbf{x}^\text{video}, \mathbf{x}^q) :\approx f_\phi(\mathbf{x}^\text{video}),
\label{eq:map-code}
\\
p(\mathbf{y} \mid \mathbf{x}^\text{video}, \mathbf{x}^q) :\approx p_\theta(\mathbf{y} \mid \mathbf{c}^*, \mathbf{x}^q),
\label{eq:map-response}
\end{gather}
where $f_\phi$ is a Spatial Encoder that encodes pertinent information from input videos as explained in \S\ref{ssec:encoder}, and $p_\theta$ is the sampling distribution from an LLM that performs reasoning on discrete tokens of spatial code and natural language. 

\subsection{Encoding Videos into Spatial Code}
\label{ssec:encoder}

We propose a \textbf{Spatial Encoder} module $f_\phi$ with neural parameters $\phi$, that predicts spatial code from an input video $\mathbf{x}^\text{video}$ (\cref{eq:map-code}), whose output $\mathbf{c} = f_\phi(\mathbf{x}^\text{video})$ has the form $\mathbf{c} = \{\mathbf{c}_i\}_{i=1}^{n}$, where each $i$ typically corresponds to an object in a scene, with object count $n$ varies across scenes.
Each code $\mathbf{c}_i = (l_i, \mathbf{p}_i, \mathbf{s}_i, \mathbf{r}_i)$ consists of a semantic label string $l_i$, position $\mathbf{p}_i \in \mathbb{R}^3$, size $\mathbf{s}_i  \in \mathbb{R}^3$, and orientation (quarternion) $\mathbf{r}_i \in \mathbb{R}^4$. A structured scene caption is inserted at the beginning of the code with global context and attributes information.

\vspace{-1.75mm}
\paragraph{Scene Captioning.}
We first employ a MLLM as a video captioner to generate structured, scene-level captions. The extracted captions include (i) global scene context, (ii) object-level descriptions, and (iii) neighboring objects descriptions. These structured captions compose the initial part of spatial code, providing rich semantic information including object attributes and contextual interactions.

\vspace{-1.75mm}
\paragraph{Feature Encoders.}
We adopt a dual-encoder design. For each input video frame with frame index $t$, we use the image encoder from SAM-2~\citep{ravi2024sam} to extract semantic feature $F^t_{\text{SAM}}$, and use the encoder from Depth Anything 3~\citep{lin2025depth} to produce  3D-aware feature $F^t_{\text{DA}}$.
These features are fused via a sequence of cross-attention layers $f_{\text{CA}}(\cdot)$. 
Then a lightweight transformer-based tracker $f_{\text{track}}(\cdot)$ from SAM-2~\citep{ravi2024sam} processes the per-frame features across frames to maintain object identity:
\begin{equation}
F^t_\text{CA} = f_{\text{CA}}(F^t_{\text{SAM}}, F^t_{\text{DA}}), \quad
F^t = f_{\text{track}}(F^t_\text{CA}, F^{t-1}).
\label{eq:encoder}
\end{equation}

\vspace{-1.75mm}
\paragraph{3D Detection Head.}
As in SAM-2, we provide a 2D bounding box prompt $\hat b_i$ with label $l_i$ on the first frame of the video clips. With the temporal-spatial feature $F^t$ from \cref{eq:encoder}, position $\mathbf{p}^t_i$, size $\mathbf{s}^t_i$, and orientation $\mathbf{r}^t_i$ for the $t$-th frame and the $i$-th object are predicted via a learnable 3D detection head $h_\phi^{\text{det}}(\cdot)$.  As ~\cite{brazil2023omni3d}, an additional 2D bounding box $b_{i}^t$ is predicted at each time stamp to stabilize training. Additionally, objects may enter or exit the field of view in the video. We predict an appearance probability $\tilde p_i^t$ for each object at frame $t$ to indicate whether it is visible. Furthermore, this process is conditioned on depth feature $h_\phi^{\text{dep}}(F^{t}_{\text{DA}})$ defined in \cref{eq:depth-head}. Overall, the 3D detection head can be written as:
\begin{equation}
(\mathbf{p}^t_i, \mathbf{s}^t_i, \mathbf{r}^t_i, b_{i}^t,\tilde p_i^t) = h_\phi^{\text{det}}(F^t, h_\phi^{\text{dep}}(F^{t}_{\text{DA}}),\hat b_i).
\end{equation}
Frame-wise predictions of the same object are merged at the scene-level based on 3D positions, forming the final spatial code $\mathbf{c}_i$ for each object (see \S\ref{sec:app-det-fusion} for more details).

\vspace{-1.5mm}
\paragraph{Depth Head.}
The supervision from 3D object detection for the 3D head is inherently sparse, providing regression targets only at the object level. Consequently, most regions in the scene, particularly background areas, contain limited geometric cues for learning robust features. 

To address this, we adopt a depth head $h_\phi^{\text{dep}}(\cdot)$ that predicts dense and informative depth features. Subsequently, a lightweight decoder $h_\phi^\text{dep-out}$ transforms these enriched features into explicit depth maps:
\begin{equation}
\label{eq:depth-head}
D_t = h_\phi^\text{dep-out} \circ h_\phi^{\text{dep}}(F^{t}_{DA}).
\end{equation}
At the same time, following~\cite{wang2025vggt}, camera parameters are also predicted by this head for supervision needs. The benefits of leveraging dense geometric supervision include (1) stabilizing learning in otherwise information-sparse regions, and (2) capturing fine-grained geometric relationships among objects at the pixel level. The benefits are empirically validated in \S\ref{ssec:exp-ablation}.

\vspace{-1.75mm}
\paragraph{Training Objective.} 
Following~\cite{brazil2023omni3d}, the spatial encoder is trained end-to-end by a multi-task loss:
\begin{equation}
\begin{aligned}
    \mathcal{L}_\text{spatial} =& \underbrace{\mathcal{L}_{\text{2D-det}}+\mathcal{L}_{\text{pos}}+\mathcal{L}_{\text{size}}+\mathcal{L}_{\text{ori}}+\mathcal{L}_{\text{chamfer}}}_{\mathcal{L}_{\text{detection}}} \\
    &+\underbrace{\mathcal{L}_{\text{camera}}+\mathcal{L}_{\text{depth}}}_{\mathcal{L}_{\text{geometry}}}+\mathcal{L}_{\text{tracking}}.
\end{aligned}
\end{equation}
The detection loss $\mathcal{L}_{\text{detection}}$ supervises frame-wise 3D bounding box predictions through multiple components: $\mathcal{L}_{\text{2D-det}}$ combines GIoU~\cite{rezatofighi2019generalized} and L1 losses for 2D box alignment; $\mathcal{L}_{\text{pos}}$ enforces 3D position accuracy using L1 loss for projected 2D centers and Laplacian aleatoric uncertainty loss~\cite{kendall2017uncertainties} for depth; $\mathcal{L}_{\text{size}}$ applies L1 loss to predicted object dimensions; $\mathcal{L}_{\text{ori}}$ supervises normalized quaternion representations with L1 loss; and $\mathcal{L}_{\text{chamfer}}$ ensures bidirectional corner-wise alignment between predicted and ground-truth 3D boxes.

The geometry loss $\mathcal{L}_{\text{geometry}}$ supervises dense spatial understanding: $\mathcal{L}_{\text{depth}}$ employs a scale-invariant loss with aleatoric uncertainty weighting and gradient regularization for pixel-wise depth prediction, while $\mathcal{L}_{\text{camera}}$ supervises camera parameter estimation through L1 losses.

Finally, $\mathcal{L}_{\text{tracking}}$ supervises the model's object appearance prediction across frames, implemented as a binary classification loss to determine whether an object is present in a given frame. More details are presented in \S\ref{sec:app-det-loss}.

\subsection{Prompting LLMs with Spatial Code}
\label{ssec:prompting}
The spatial code $\mathbf{c}$ predicted by the Spatial Encoder from \S\ref{ssec:encoder} enables text-only LLMs to perform explicit coordinate-based reasoning on input videos. The inference procedure follows \cref{eq:map-response} and is expanded below.

Let $\mathbf{c}_i$ be the $i$-th object's spatial code from the Spatial Encoder's prediction $\mathbf{c}$. It is serialized into text:
\begin{equation}
\texttt{str}(\mathbf{c}_i) = \texttt{"\{"}\texttt{bbox\_3d}: [\mathbf{p}_i, \mathbf{s}_i, \mathbf{r}_i], \texttt{label}: l_i\texttt{"\}"},
\end{equation}
and the LLM outputs response $\mathbf{y}$ autoregressively via
\begin{equation}
p_\theta(\mathbf{y} \mid \mathbf{c}, \mathbf{x}^q) = \prod_{j=1}^{L} p_\theta(\mathbf{y}_j \mid \mathbf{y}_{<j}, \texttt{str}(\mathbf{c}), \mathbf{x}^q),
\end{equation}
where $L$ is the length of response with response tokens $\mathbf{y}_j$, and $\texttt{str}(\mathbf{c}) = \bigoplus_{i=1}^{n} \texttt{str}(\mathbf{c}_i)$ concatenates all spatial codes.
The code $\mathbf{c}$ is immediately interpretable by pre-trained LLM without finetuning, improving the LLM's performance on spatial queries (\eg~``Can the lamp fit to the right of the table?'') through explicit coordinate reasoning.

\subsection{Reinforcement Learning with Spatial Rubric Reward}
\label{ssec:rl}
To further improve the LLM reasoning policy, we employ RL finetuning. While standard RL with verifiable rewards~\citep{lambert2024tulu} relies solely on outcome-based verification, we augment it with domain-specific process supervision inspired by prior work on step-level rewards~\citep{uesato2022solving,lightman2023lets}. 

Our reward function combines the  verifiable outcome (accuracy) reward with rule-based spatial rubrics that evaluate intermediate reasoning quality:
\begin{equation}
r(\mathbf{y}|\mathbf{x}) = \underbrace{r_{\text{acc}}(\mathbf{y}, a^*)}_{\text{accuracy}} + \underbrace{r_{\text{format}}(\mathbf{y})}_{\text{format compliance}} + \underbrace{r_{\text{rubric}}(\mathbf{y}, \mathbf{x})}_{\text{spatial rubrics}},
\label{eq:reward}
\end{equation}
where $a^*$ is the ground-truth answer, $\mathbf{y}$ is a model response, and $\mathbf{x}$ is the input context. For the first term, $r_{\text{acc}}=1$ if the answer extracted from response $\mathbf{y}$ matches $a^*$, and $0$ otherwise. The format compliance term $r_{\text{format}}$ rewards responses following the expected structure (\eg~concluding with ``Final Answer: [A/B/C/D]'') and penalizes degenerate repetition patterns. The last term is explained below.

\begin{figure}[ht!]
\vspace{-1.75mm}
  \centering
   \includegraphics[width=0.85\linewidth]{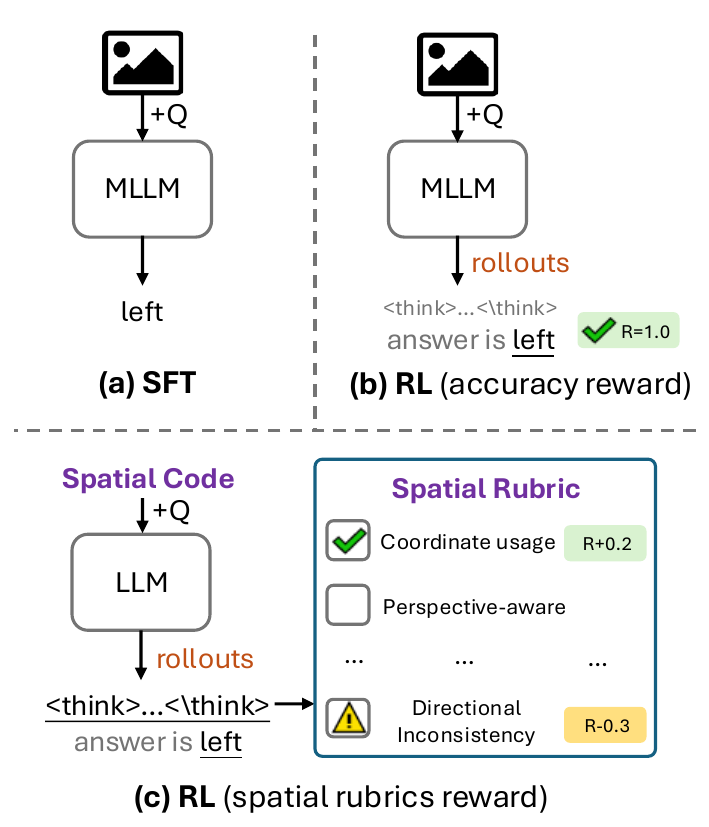} 
      \vspace{-1.5mm}
  \caption{\textbf{Comparison of Spatial Rubric Reward (c) against conventional SFT (a) and RL (b).} Unlike traditional methods, our framework utilizes the 3D spatial codes as primary input. Applying a structured spatial rubric reward to model rollouts significantly improves the quality of spatial reasoning.
  }
  \label{fig:spatialrubrics}
   \vspace{-3mm}
\end{figure}

\vspace{-1.75mm}
\paragraph{Spatial Rubric Reward.}
The spatial rubric term $r_{\text{rubric}}$ in \cref{eq:reward} evaluates reasoning quality through a weighted sum of binary indicator functions:
\begin{equation}
r_{\text{rubric}}(\mathbf{y}, \mathbf{x}) = \sum_{i=1}^{K} w_i \cdot \psi_i(\mathbf{y}, \mathbf{x}),
\end{equation}
where $\psi_i: (\mathbf{y}, \mathbf{x}) \to \{0, 1\}$ detects whether the response $\mathbf{y}$ exhibits reasoning pattern $i$, and $w_i \in \mathbb{R}$ assigns positive weights to desired behaviors and negative weights to failure modes. The indicators $\{\psi_i\}$ are task-specific and target common spatial reasoning errors: (1) \emph{world-coordinate confusion} --- using global axes instead of object-centric coordinates; (2) \emph{missing coordinate transformation} --- skipping local reference frame construction; (3) \emph{reasoning-answer inconsistency} --- correct intermediate analysis but wrong final answer. For example, in the relative direction task, we reward explicit construction of local basis vectors ($w=+0.25$) and penalize direct mapping from world coordinates ($w=-0.25$). The total reward is clipped to $[-0.5, 1.8]$. Details are provided in \S\ref{app:rubric}.

\vspace{-1.75mm}
\paragraph{Training Objective.}
We finetune pre-trained LLMs using GRPO~\citep{shao2024deepseekmath}, which computes advantages by comparing multiple sampled responses. For each prompt $\mathbf{x}$, we sample $G$ responses $\{\mathbf{y}^{(i)}\}_{i=1}^G$ and compute group-normalized advantages:
\begin{equation}
A^{(i)} = \frac{r(\mathbf{y}^{(i)}|\mathbf{x}) - \mu_G}{\sigma_G + \epsilon},
\end{equation}
where $\mu_G$ and $\sigma_G$ are the mean and standard deviation of rewards within the group, and $\epsilon$ is a small constant for numerical stability.
The policy $\pi_{\theta}$is optimized via:
\begin{equation}
\mathcal{L}_{\text{GRPO}}(\theta) = \mathbb{E}\left[\sum_{i=1}^{G} A^{(i)} \log \pi_\theta(\mathbf{y}^{(i)}|\mathbf{x})\right] - \beta \cdot D_{\text{KL}}[\pi_\theta \| \pi_{\text{ref}}],
\end{equation}
where $\beta$ controls the KL penalty against a reference policy $\pi_{\text{ref}}$, which is the frozen Qwen3-4B~\cite{yang2025qwen3}.

\begin{table*}[ht!]
\centering
\caption{\textbf{Video spatial reasoning results on VSI-Bench and Video-RoboSpatial.} \fullname{} achieves state-of-the-art performance, outperforming proprietary MLLMs (\eg~GPT-5o) and spatial-centric methods. Gray rows indicate fair comparisons.
} 
\vspace{-1mm}
\renewcommand{\arraystretch}{1.2}
\resizebox{\linewidth}{!}{
\begin{tabular}{l c |>{\columncolor{blue!5}}c c c c c c c c c | c}
\toprule
\multirow{2}{*}{\textbf{Methods}} & \multirow{2}{*}{\textbf{Size}} & \multicolumn{9}{c|}{\textbf{VSI-Bench}~\citep{yang2025thinking}} & \textbf{Video-RoboSpatial} \\
\cmidrule(lr){3-11} \cmidrule(lr){12-12}
 & & \textbf{Avg.} & \makecell{\textbf{Obj.}\\\textbf{Count}} & \makecell{\textbf{Abs.}\\\textbf{Dist.}} & \makecell{\textbf{Obj.}\\\textbf{Size}} & \makecell{\textbf{Room}\\\textbf{Size}} & \makecell{\textbf{Rel.}\\\textbf{Dist.}} & \makecell{\textbf{Rel.}\\\textbf{Dir.}} & \makecell{\textbf{Route}\\\textbf{Plan}} & \makecell{\textbf{Appear.}\\\textbf{Order}} & \textbf{Config.} \\
\midrule
Human Level & -- & 79.2 & 84.3 & 47.0 & 60.4 & 45.9 & 94.7 & 95.8 & 95.8 & 100 & 92.1 \\
\midrule
\multicolumn{12}{l}{\textit{Proprietary MLLMs}} \\
Seed-1.6 & 230B & 49.9 & 43.5 & 34.3 & 66.1 & \underline{52.8} & 55.0 & 35.7 & 44.3 & 67.9 & 54.3 \\
Gemini-1.5 Pro & -- & 48.8 & 49.6 & 30.9 & 64.1 & 49.4 & 51.3 & 48.1 & 42.0 & 68.0 & -- \\
Gemini-2.5-Pro & -- & 53.5 & 46.0 & 37.3 & \underline{68.7} & \textbf{54.3} & 61.9 & 43.9 & \underline{47.4} & 68.7 & 53.3 \\
GPT-4o & -- & 34.0 & 46.2 & 5.3 & 43.8 & 38.2 & 37.0 & 41.3 & 31.5 & 28.5 & 53.0 \\
GPT-5 & -- & 55.0 & 53.3 & 34.4 & \textbf{73.3} & 47.5 & \underline{63.7} & 48.6 & \textbf{50.2} & \textbf{68.9} & 60.3 \\
\midrule
\multicolumn{12}{l}{\textit{Spatial-centric MLLMs}} \\
SpatialLadder~\citep{spatialladder} & 3B & 44.8 & 62.1 & 35.3 & 61.9 & 41.4 & 45.6 & 46.4 & 27.3 & 38.5 & 49.3 \\
Spatial-MLLM~\citep{spatialmllm} & 4B & 46.3 & \underline{66.6} & 38.0 & 63.6 & 35.4 & 40.4 & 48.2 & 32.9 & 44.3 & 49.0 \\
SpaceR~\citep{spacer} & 7B & 41.5 & 44.5 & 24.7 & 53.5 & 37.3 & 41.9 & 46.1 & 29.3 & 54.8 & 56.0 \\
\midrule
\multicolumn{12}{l}{\textit{Open-source MLLMs}} \\
LLaVA-Video~\citep{lin2024video} & 7B & 35.6 & 48.5 & 14.0 & 47.8 & 24.2 & 43.5 & 42.4 & 34.0 & 30.6 & 52.0 \\
LLaVA-Video~\citep{lin2024video} & 72B & 40.9 & 48.9 & 22.8 & 57.4 & 35.3 & 42.4 & 36.7 & 35.0 & 48.6 & 58.0 \\
LLaVA-OneVision~\citep{li2025llava} & 7B & 32.4 & 47.7 & 20.2 & 47.4 & 12.3 & 42.5 & 35.2 & 29.4 & 24.4 & 54.3 \\
LLaVA-OneVision~\citep{li2025llava} & 72B & 40.2 & 43.5 & 23.9 & 57.6 & 37.5 & 42.5 & 39.9 & 32.5 & 44.6 & 56.0 \\
Qwen2.5-VL~\citep{qwen25vl} & 7B & 32.3 & 32.8 & 18.1 & 43.8 & 31.7 & 38.0 & 37.4 & 28.3 & 27.9 & 49.7 \\
\textcolor{gray}{Qwen3-VL~\citep{yang2025qwen3}} & \textcolor{gray}{8B} & \textcolor{gray}{55.0} & 
\textcolor{gray}{52.1} & \textcolor{gray}{44.7} & \textcolor{gray}{60.4} & \textcolor{gray}{43.1} & \textcolor{gray}{56.6} & \textcolor{gray}{56.3} & \textcolor{gray}{38.1} & \textcolor{gray}{70.2} & \textcolor{gray}{48.0} \\
\rowcolor{gray!10} Qwen3-VL~\citep{yang2025qwen3} & 4B & \cellcolor{blue!5}52.8 & 53.1 & 46.3 & 63.4 & 48.0 & 53.3 & 49.9 & 37.1 & 58.9 & 52.7 \\
\rowcolor{gray!10} Qwen3-VL~\citep{yang2025qwen3} + 2D box & 4B & \cellcolor{blue!5}54.5 & 66.6 & 42.3 & 57.8 & 40.8 & 56.9 & 52.7 & 37.6 & 66.9 & 52.7 \\
\midrule
\rowcolor{gray!10} \fullname{} (w/o RL) & 4B & \cellcolor{blue!5}54.6 & 58.3 & 37.0 & 71.2 & 52.4 & 54.2 & 50.2 & 35.6 & 63.9 & 65.3 \\
\rowcolor{gray!10} \fullname{} (w/o RL) + 2D box & 4B & \cellcolor{blue!5}56.5 & \textbf{95.2} & \underline{50.0} & 50.9 & 19.9 & \textbf{63.7} & 55.5 & 25.3 & 58.4 & \underline{65.3} \\
\rowcolor{gray!10} \textbf{\fullname{}} & 4B & \cellcolor{blue!5}\underline{57.0} & 58.3 & 39.0 & \underline{73.0} & 52.4 & 57.8 & \underline{55.9} & 38.7 & 63.9 & 67.0  \\
\rowcolor{gray!10} \textbf{\fullname{}} + 2D box & 4B & \cellcolor{blue!5}\textbf{60.0} & \textbf{95.2} & \textbf{60.7} & 50.8 & 33.1 & 62.0 & \textbf{87.1} & 32.5 & 59.0 & \textbf{67.0} \\
\bottomrule
\end{tabular}
}
\label{tab:main}
\vspace{-4mm}
\end{table*}

\section{Experiments}
\label{sec:experiments}

We evaluate the framework on video spatial reasoning (\S\ref{ssec:exp-reasoning}) and video 3D perception (\S\ref{ssec:exp-perception}) benchmarks, followed by ablation studies (\S\ref{ssec:exp-ablation}) and analysis (\S\ref{ssec:exp-analysis}).

\subsection{Spatial Reasoning from Video Inputs}
\label{ssec:exp-reasoning}

The task requires model to reason about object attributes, relations, and dynamics in space and time from video inputs.

\noindent\textbf{Benchmarks.}
We evaluate on \textbf{VSI-Bench}~\citep{yang2025thinking}, which provides evaluation across: object counting, absolute/relative distance, object/room size, relative direction, route planning, and appearance order. The input for baseline models is RGB videos, and we require the model to output answers with reasoning.
We further introduce \textbf{Video-RoboSpatial}, a benchmark for spatial reasoning over continuous video streams with annotated 6D object states, with particular emphasis on precise 3D orientation. Built upon ARKitScenes~\citep{baruch2021arkitscenes}, it extends RoboSpatial~\citep{liang2024robospatial} question templates to video. 
For ambiguity avoidance, we by default give the model explicit 2D bounding boxes for identifying objects of interests.

\noindent\textbf{Baselines.}
We compare against: (1) \emph{proprietary MLLMs}: GPT-4o~\cite{hurst2024gpt}, GPT-5o, Gemini-1.5/2.5-Pro~\citep{comanici2025gemini}, Grok-4, and Seed-1.6; (2) \emph{spatial-centric VLMs}: SpatialLadder~\citep{spatialladder}, Spatial-MLLM~\citep{spatialmllm}, and SpaceR~\citep{spacer}; (3) \emph{open-source VLMs}: LLaVA-Video~\citep{lin2024video}, LLaVA-OneVision~\citep{li2025llava}, and Qwen-VL series~\citep{qwen25vl,yang2025qwen3}.

\noindent\textbf{Results.}
As shown in \cref{tab:main}, even without RL training, \fullname{} improves over the Qwen3-VL-4B baseline by 1.6\%. When additionally provided with 2D bounding box annotations, performance reaches \textbf{56.5\%}, surpassing GPT-5o (55.0\%), Gemini-2.5-Pro (53.5\%), and Qwen3-VL-8B (55.0\%). Compared to the non-RL counterparts, incorporating the spatial rubric reward yields consistent gains of \textbf{+3.4\%} (without 2D boxes) and \textbf{+3.5\%} (with 2D boxes), demonstrating the effectiveness of RL training with perspective-aware spatial reasoning supervision. On Video-RoboSpatial, \fullname{} achieves \textbf{67.0\%} on configuration reasoning, outperforming the second-best method (GPT-5) by \textbf{6.7\%}.

\subsection{3D Perception from Video Inputs}
\label{ssec:exp-perception}
We evaluate the perception capability of the Spatial Encoder, comparing it with several recent SOTA perception models.

\begin{table}[!ht]
\centering
\small
\caption{\textbf{Video 3D perception performance comparison.} Our Spatial Encoder achieves state-of-the-art performance on scene-level F1, demonstrating strong spatial-temporal consistency. Image-based detectors operate on single frames and to support video input we aggregate predictions across frames. Point-cloud-based detectors such as SceneScript require ground-truth point clouds as input (video support is not yet open-sourced). MLLMs process videos as sequences of image frames. No existing open-source method supports video 3D perception; our approach fills this gap.}
\renewcommand{\arraystretch}{1.2}
\resizebox{0.49\textwidth}{!}{
\begin{tabular}{cl cc cc}
\toprule
\multirow{2}{*}{\textbf{Input}}& \multirow{2}{*}{\textbf{Methods}} & \multicolumn{2}{c}{\textbf{ARKitScenes}} & \multicolumn{2}{c}{\textbf{ScanNet}} \\
\cmidrule(lr){3-4} \cmidrule(lr){5-6}
\cmidrule(lr){3-4} \cmidrule(lr){5-6}
& & F1@.25 & F1@.5 & F1@.25 & F1@.5 \\
\midrule
\multirow{2}{*}{\makecell{Image \\detector} } 
& 3D-MOOD{\scriptsize ~\citep{yang20253d}} & 0.066 & 0.009 & 0.083 & 0.025 \\
& DetAny3D{\scriptsize ~\citep{zhang2025detect}} & 0.094 & 0.012 & 0.119 & 0.024 \\
\midrule
\multirow{2}{*}{\makecell{Point-cloud \\detector}} 
& SpatialLM{\scriptsize ~\citep{mao2025spatiallm}} & \underline{0.122} & \underline{0.055} & \underline{0.134} & 0.025 \\
& SceneScript{\scriptsize ~\citep{avetisyan2024scenescript}} & 0.020 & 0.014 & 0.101 & \underline{0.043} \\
\midrule
\multirow{2}{*}{MLLMs} 
& Qwen3-VL-4B{\scriptsize ~\citep{yang2025qwen3}} & 0.041 & 0.011 & 0.077 & 0.023 \\
& Qwen3-VL-235B{\scriptsize ~\citep{yang2025qwen3}} & 0.034 & 0.008 & 0.087 & 0.028 \\
\midrule
\rowcolor{gray!10} \textbf{{\makecell{Video \\ detector }} } & \textbf{Spatial Encoder (Ours)} & \textbf{0.156} & \textbf{0.082}& \textbf{0.209} & \textbf{0.062} \\
\bottomrule
\end{tabular}
}
\vspace{-4mm}
\label{tab:perception}
\end{table}

\paragraph{Setup.}
Evaluation is conducted on the test set of ARKitScenes~\citep{baruch2021arkitscenes} (549 scenes) and the VSI-bench subset of ScanNet~\citep{dai2017scannet} (88 scenes) using scene-wise F1 scores at IoU thresholds of 0.25 (F1@.25) and 0.5 (F1@.5). For ARKitScenes, we use oriented bounding boxes for IoU calculation. For ScanNet, we use axis-aligned IoU due to inconsistent orientation annotations in the dataset.
We compare against: (1) \emph{image-based 3D detectors}: 3D-MOOD~\citep{yang20253d} and DetAny3D~\citep{zhang2025detect}; (2) \emph{point-cloud methods}: SpatialLM~\citep{mao2025spatiallm} and SceneScript~\citep{avetisyan2024scenescript}; (3) \emph{MLMMs}: Qwen3-VL~\citep{yang2025qwen3}. For methods with frame-wise predictions, we aggregate all predictions at the scene level by Non-Maximum Suppression. Bounding boxes are matched with the ground truth using the Hungarian algorithm for evaluation.

\vspace{-2.75mm}
\paragraph{Results.}
As shown in \cref{tab:perception}, our Spatial Encoder achieves state-of-the-art results across both datasets. On ARKitScenes, we achieve F1@0.25 of 0.156 scene-wise, surpassing previous image-based methods by at least 0.06. On ScanNet scene-wise evaluation, we reach F1@0.25 of 0.209, significantly outperforming DetAny3D (0.119) and Qwen3-VL-235B (0.087). Notably, despite relying solely on video inputs, our method surpasses point-cloud-based approaches that operate on explicit 3D geometry on both datasets. These results demonstrate that our spatial encoder effectively maintains spatial-temporal consistency and produces more accurate 3D spatial codes.

\subsection{Ablation Studies}
\label{ssec:exp-ablation}

\begin{table}[t]
\centering
\small
\caption{\textbf{Ablation on encoder design.} Dep. Err.: Depth Error. Scale Err.: Scale Error.}
\vspace{-1mm}
\renewcommand{\arraystretch}{1.}
\resizebox{\linewidth}{!}{
\begin{tabular}{ccc cccc}
\toprule
SAM2 Enc. & DA3 Enc. & Depth Head & F1@.25 & 3D IoU$\uparrow$ & Dep. Err.$\downarrow$ & Scale Err.$\downarrow$ \\
\midrule
\checkmark & - & - & 0.27& 0.10 & 0.50 & 0.63 \\
\checkmark & \checkmark & - & 0.31 &0.12 & 0.42 & 0.57 \\
\rowcolor{gray!10} \checkmark & \checkmark & \checkmark & \textbf{0.32} & \textbf{0.14} & \textbf{0.31} & \textbf{0.55} \\
\bottomrule
\end{tabular}
}
\vspace{-3mm}
\label{tab:ablation_encoder}
\end{table}

\begin{table}[t]
\centering
\small
\caption{\textbf{Ablation on RL reward.} Spatial rubric rewards yield large gains on direction and distance tasks.}
\renewcommand{\arraystretch}{1.2}
\resizebox{0.9\linewidth}{!}{
\begin{tabular}{l cccc}
\toprule
\textbf{Training} & \textbf{Avg.} & \textbf{Rel. Dir.} & \textbf{Abs. Dist.} & \textbf{Config.} \\
\midrule
w/o spatial code & 51.8 & 48.8  & 42.6 & 47.0 \\
w/o RL & 56.5 & 55.5 & 50.0 & 65.3 \\
 w/ Accuracy Reward & 57.6 & 59.3 & 52.3 & 63.0 \\
\rowcolor{gray!10} w/ Spatial Rubric & \textbf{60.0} & \textbf{87.1} & \textbf{60.7} &  \textbf{67.0} \\
\bottomrule
\end{tabular}
}
\vspace{-2mm}
\label{tab:ablation_rl}
\end{table}

\paragraph{Spatial Encoder Design.}
\cref{tab:ablation_encoder} ablates the encoder architecture. We evaluate localization accuracy by F1@.25, 3D IoU, metric depth error, and scale error. Using only SAM-2 yields poor F1@.25 at 0.27 and large depth error at 0.5. Incrementally adding depth loss and depth head improves performance, achieving the highest F1@.25 (+0.05) and lowest depth error (-0.19).

\vspace{1.75mm}
\paragraph{Spatial Rubric Reward.}
\cref{tab:ablation_rl} compares training strategies. The spatial rubric reward achieves the largest gains on direction-sensitive tasks (+31.6\% on Rel. Dir.), confirming it encourages perspective-aware reasoning.

\subsection{Key Finding: Perception Quality Bounds Reasoning}
\label{ssec:exp-analysis}

A central finding work is that \textbf{spatial reasoning performance is largely determined by 3D perception quality}, not LLM capacity alone. Evidence is provided below.

\begin{table}[ht]
\centering
\small
\caption{\textbf{GT vs. predicted spatial codes.} The same LLM achieves 73.2\% reasoning accuracy on VSI-Bench with perfect perception.
} 
\vspace{-2mm}
\renewcommand{\arraystretch}{1.2}
\resizebox{0.4\textwidth}{!}{
\begin{tabular}{l ccc}
\toprule
\textbf{Spatial Code Source} & \textbf{RL} & \textbf{Perception} & \textbf{Reasoning} \\
\midrule
Ground-truth codes & & 1.00 & \textbf{68.9} \\
Spatial Encoder (ours) & & 0.52 & 56.5 \\
\midrule
Ground-truth codes & $\checkmark$ & 1.00 & \textbf{73.2} \\
Spatial Encoder (ours) & $\checkmark$ & 0.52 & 60.0 \\
\textit{Human performance} & & -- & 79.2 \\
\bottomrule
\end{tabular}
}
\vspace{-3mm}
\label{tab:gt_vs_pred}
\end{table}

\paragraph{Ground-Truth (GT) vs. Predicted Spatial Codes.}
Table~\ref{tab:gt_vs_pred} shows that with GT spatial codes, the same 4B LLM achieves \textbf{72.3\%} accuracy, narrowing the human-machine gap. With predicted codes from our Spatial Encoder (F1@0.25$\approx$0.52), accuracy drops to 60.0\%. This 12.3\% gap reflects perception errors propagating into reasoning.

\paragraph{Model Scale vs. Representation Quality.}
\cref{fig:teaser} visualizes that MLLMs processing raw video (GPT-5o, Gemini-2.5, Qwen3-VL, Seed-1.6) plateau at 50--55\% regardless of scale (4B to 230B parameters). In contrast, our 4B model with explicit spatial codes achieves 60.0\%—and 73.2\% with GT codes. This 10\%$+$ gap between spatial-code reasoning and the best MLLMs demonstrates that \textbf{representation quality, not parameter count, is the limiting factor}.

\vspace{-1mm}
\paragraph{Implication.}
Recall the factorization in \cref{eq:factorization}.
These findings confirm that errors in the perception model $p(\mathbf{c} \mid \mathbf{x}^\text{video})$ propagate directly to reasoning, indicating that 3D perception remains a  bottleneck for spatial reasoning.

\subsection{Implementation Details}
\label{ssec:exp-details}

\paragraph{Spatial Encoder.}

The training dataset  includes: CA-1M~\cite{lazarow2025cubify} (439K objects, 200M frames), Hyperism~\cite{roberts2021hypersim} (40K objects, 60K frames), and Aria Digital Twin~\cite{pan2023aria} (24K objects, 254K frames). Both encoders are frozen, and the prediction heads are trained on 4$\times$ A100 GPUs with a batch size of 4 for 800k steps (about 10 days). We use AdamW optimizer and cosine scheduler with a initial learning rate of 1e-4. The ablation study is conducted on a subset of CA-1M. More details are provided in \S\ref{sec:app-det-details}.

Additionally, we provide a training-free implementation of the spatial encoder. Bounding boxes are acquired by lifting SAM2 predictions into 3D and clustering based on lifted points. The implementation details are provided in \S\ref{sec:app-encoder-v2}.

\vspace{-2mm}

\paragraph{Reinforcement Learning.}
Qwen3-4B~\citep{yang2025qwen3} with GRPO, number of rollouts $G=16$, $\beta=0.01$, lr 1e-6, 4$\times$ A100 GPUs, 1 day. Details are provided in \S\ref{app:hyperparams}.

\subsection{Qualitative Analysis}
\label{ssec:exp-qualitative}

\begin{figure*}[!t]
\centering
\includegraphics[width=\textwidth]{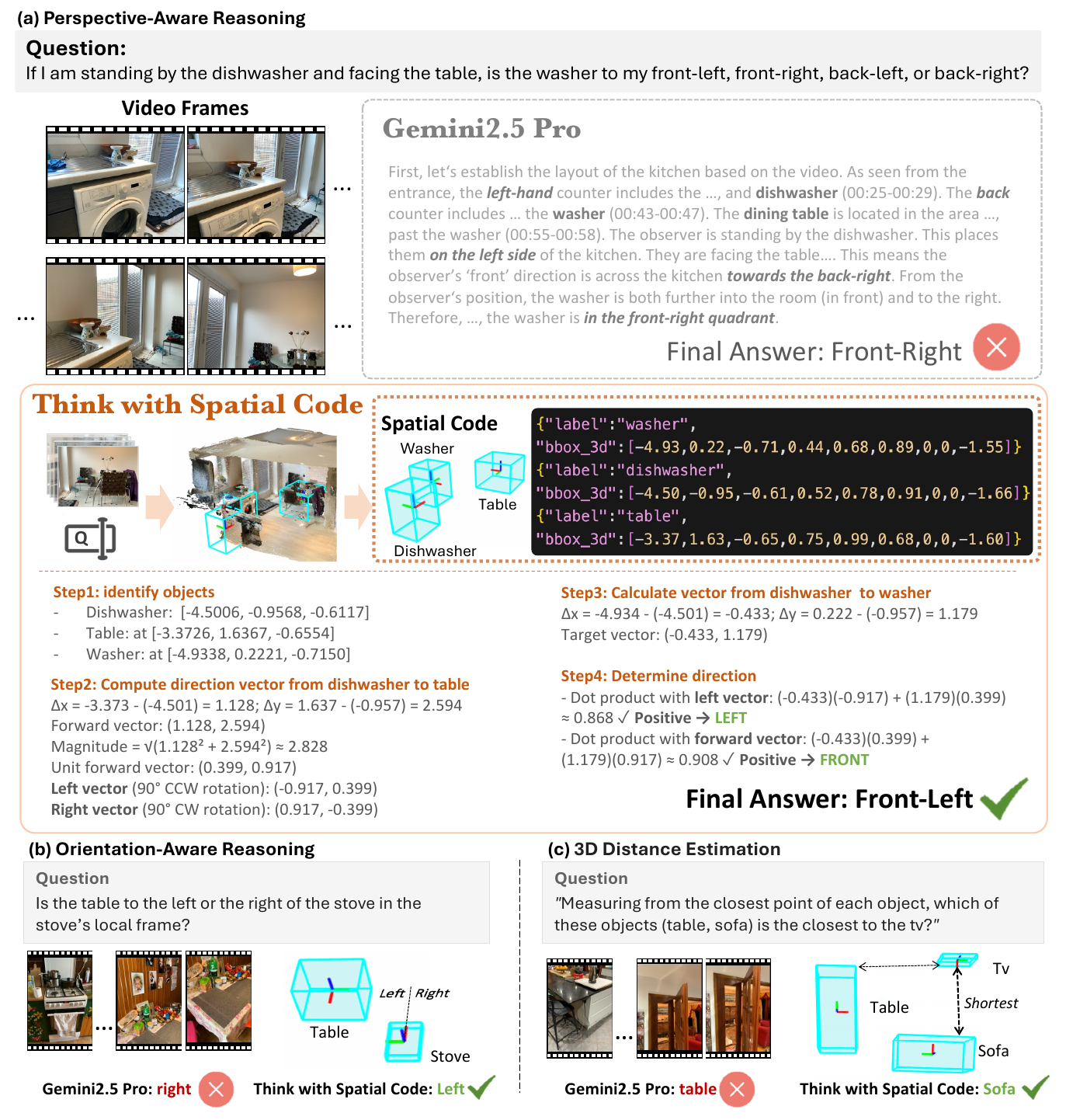}
\vspace{-3.5mm}
\caption{\textbf{Qualitative comparison.} 
We show three examples where \fullname{} succeeds while Gemini 2.5 Pro fails.
(a) \textbf{Perspective-aware reasoning with detailed reasoning trace:} The question requires model to reason from a specific observer viewpoint. Video-based models confuse absolute positions with observer-relative directions, while our spatial codes enable step-by-step coordinate transformation with precise calculation and significantly improves reasoning accuracy.
(b)\textbf{ Orientation-aware reasoning:} The question requires understanding object orientation (yaw angles). MLLMs rely on visual appearance, while our spatial codes provide explicit orientation parameters for accurate inference.
(c) \textbf{3D distance estimation:} The task requires metric depth measurements. MLLMs use ambiguous 2D visual cues, while our spatial codes provide precise 3D coordinates for reliable distance calculation.
}
\vspace{-1.5mm}
\label{fig:qualitative}
\end{figure*}

Qualitative comparisons in Fig.~\ref{fig:qualitative} highlight three common failure modes of MLLMs that our method addresses: (1) perspective-aware reasoning (Fig.~\ref{fig:qualitative}a), (2) orientation-aware reasoning (Fig.~\ref{fig:qualitative}b), and (3) 3D distance estimation (Fig.~\ref{fig:qualitative}c). These questions requires model to understand perspective change, object orientation, and metric distance, respectively. These are tasks that current MLLMs struggle with due to reliance on ambiguous 2D visual cues. In contrast, our spatial codes provide explicit 3D cues, which enables precise calculation and suppresses hallucination.

The examples illustrate why \textit{explicit spatial representations} outperform end-to-end MLLMs: they eliminate visual ambiguities and provide the LLM with precise geometric information for reasoning.

\section{Related Work}

\textbf{Vision-Language Models.}
The development of VLMs has progressed from contrastive approaches like CLIP~\citep{radford2021clip, chen2024vitamin} to MLLMs models~\citep{alayrac2022flamingo, li2023blip2, liu2024llava,chen2024efficient,li2025llava}. Recent efforts extend to video understanding through Video-LLaVA~\citep{lin2024video} and proprietary systems such as Gemini~\citep{comanici2025gemini} and Qwen-VL~\citep{qwen25vl,yang2025qwen3}. However, these models rely on 2D appearance features and lack explicit 3D geometric understanding, struggling with tasks requiring precise spatial localization or perspective-aware reasoning. Our work addresses this gap by augmenting VLMs with structured 3D spatial codes that provide geometric grounding.

\textbf{Spatial Reasoning.}
To evaluate and improve spatial understanding, benchmarks have evolved from synthetic datasets like CLEVR~\citep{johnson2017clevr} and Spatial457~\cite{wang2025spatial457} to realistic settings including SpatialBench~\citep{cai2024spatialbench}, RoboSpatial~\citep{liang2024robospatial}, and video-based VSI-Bench~\citep{yang2025thinking}. Concurrent approaches such as SpaceR~\citep{spacer}, SpatialLadder~\citep{spatialladder}, and Spatial-MLLM~\citep{spatialmllm} improve spatial reasoning through reinforcement learning or progressive training. SpatialVLM~\citep{chen2024spatialvlm} grounds VLMs with metric depth for spatial QA, but operates on single images without temporal consistency. These methods either lack explicit 3D grounding or do not handle video input. In contrast, our framework parses streaming video into temporally coherent 3D spatial codes, enabling reasoning over explicit geometric representations across frames.

\textbf{Vision as Inverse Graphics.}
The paradigm of vision-as-inverse-graphics and analysis-by-synthesis~\citep{yuille2006vision} has inspired approaches from classical structure-from-motion~\citep{schonberger2016sfm} to learning-based scene understanding~\citep{wu2017neural,yao20183d}. MarrNet~\citep{marrnet} operationalizes this through a 2D--2.5D--3D pipeline. Recent advances include depth estimation models like Depth Anything~\citep{yang2024depthanything,lin2025depth}, segmentation foundations such as SAM-2~\citep{ravi2024sam}, and 3D detectors including 3D-MOOD~\citep{yang20253d} and DetAny3D~\citep{zhang2025detect}. Building on these, structured scene representations like SceneScript~\citep{avetisyan2024scenescript} and SpatialLM~\citep{mao2025spatiallm} generate symbolic 3D descriptions in natural language. However, these methods require point cloud input and are restricted to pre-defined object categories, with performance heavily dependent on sensor quality. Our approach differs fundamentally: we operate directly on RGB video with open-vocabulary detection through natural language prompts, eliminating the dependency on 3D sensors while maintaining robust spatial understanding through code-like geometric representations.
\section{Conclusion}
We introduced \fullname, a framework that transforms RGB video into explicit 3D spatial representations for visual question answering from videos. We propose a Spatial Encoder architecture that unifies semantic and geometric understanding through a dual-encoder architecture, generating structured spatial codes that bridge perception and language reasoning. 
Combined with reinforcement learning using spatial rubric rewards on top of pre-trained LLMs, our approach achieves state-of-the-art performance on VSI-Bench, outperforming both proprietary and open-source VLMs. We will release our code, models, and training recipes to facilitate future research.

\section*{Acknowledgement}
We gratefully acknowledge helpful discussions with Joshua Tenenbaum, Tianjian Li, Hao Chen, Sophia Qian, Daniel Khashabi, and Yana Wei in the early stages of this work.
\nocite{langley00}

\bibliography{main}
\bibliographystyle{icml2026}
\newpage 
\appendix

\setcounter{page}{1}
\begin{center}
{\Large\bf Supplementary Material}\\[1em]
\end{center}
\vspace{2em}



\section{Reinforcement Learning Implementation Details}
\label{app:implementation}

\subsection{Training Hyperparameters}
\label{app:hyperparams}

Table~\ref{tab:grpo_hyperparams} summarizes the GRPO training hyperparameters used in our experiments.

\begin{table}[h]
\centering
\caption{GRPO training hyperparameters.}
\renewcommand{\arraystretch}{1.2}
\label{tab:grpo_hyperparams}
\resizebox{1\columnwidth}{!}{%
\begin{tabular}{lcc}
\toprule
\textbf{Hyperparameter} & \textbf{Symbol} & \textbf{Value} \\
\midrule
\multicolumn{3}{l}{\textit{GRPO Algorithm}} \\
Group size (samples per prompt) & $G$ & 16 \\
KL penalty coefficient & $\beta$ & 0.01 \\
Sampling temperature & $\tau$ & 0.7 \\
Reward clipping range & -- & $[-0.5, 1.8]$ \\
\midrule
\multicolumn{3}{l}{\textit{Optimization}} \\
Base model & -- & Qwen3-4B-Instruct \\
Learning rate & $\eta$ & $1 \times 10^{-6}$ \\
Batch size per device & -- & 4 \\
Gradient accumulation steps & -- & 2 \\
Number of GPUs & -- & 4 \\
Effective batch size & $B$ & 32 \\
Training epochs & -- & 1 \\
\midrule
\multicolumn{3}{l}{\textit{Sequence Length}} \\
Max prompt length & $L_{\text{prompt}}$ & 4096 \\
Max completion length & $L_{\text{comp}}$ & 2048 \\
\midrule
\multicolumn{3}{l}{\textit{Infrastructure}} \\
Precision & -- & BF16 \\
Distributed strategy & -- & DeepSpeed ZeRO-2 \\
Gradient checkpointing & -- & Enabled \\
\bottomrule
\end{tabular}%
}
\end{table}

\subsection{Prompt Template}
\label{app:prompt}

Our prompt template consists of four components: (1) system instructions with coordinate conventions, (2) 3D bounding box data, (3) task-specific instructions, and (4) the question with answer format. Figure~\ref{fig:prompt_structure} illustrates the overall structure.

\begin{figure}[h]
\centering
\begin{tcolorbox}[
    colback=gray!3,
    colframe=gray!60,
    title={\small\textbf{Prompt Structure Overview}},
    fonttitle=\bfseries,
    boxrule=0.5pt,
    arc=2pt,
    width=0.85\columnwidth
]
\small
\textbf{1. System Instructions}\\
\textit{``You are a multimodal reasoning model that interprets structured scene inputs...''}\\[4pt]
\textbf{2. Coordinate System Conventions}\\
\textit{``World Frame: Z-axis points Up. X and Y are horizontal.''}\\
\textit{``Yaw: 0 rad is along +X. Positive Yaw is counter-clockwise.''}\\[4pt]
\textbf{3. Measurement Definitions}\\
\textit{``Object Size: max(x\_size, y\_size, z\_size)''}\\
\textit{``Distance: minimum Euclidean distance between closest points''}\\[4pt]
\textbf{4. 3D Bounding Box Data} (JSON format)\\[4pt]
\textbf{5. Task-Specific Instructions} (varies by task type)\\[4pt]
\textbf{6. Question + Answer Format}
\end{tcolorbox}
\caption{High-level prompt structure for spatial reasoning tasks.}
\label{fig:prompt_structure}
\end{figure}

\paragraph{Bounding Box Format.}
Each object is represented as a JSON dictionary with label and 3D bounding box parameters:
\begin{tcolorbox}[
    colback=blue!3,
    colframe=blue!30,
    boxrule=0.3pt,
    left=2pt, right=2pt, top=2pt, bottom=2pt
]
\small
\texttt{\{"label": "cabinet", "bbox\_3d": [x\_center, y\_center, z\_center,}\\
\texttt{~~x\_size, y\_size, z\_size, roll, pitch, yaw]\}}
\end{tcolorbox}
\noindent where \texttt{(x\_center, y\_center, z\_center)} is the center position, \texttt{(x\_size, y\_size, z\_size)} are dimensions, and \texttt{yaw} is the rotation around the Z-axis in radians.

\paragraph{Task-Specific Instructions.}
Table~\ref{tab:task_instructions} summarizes the task-specific instructions injected into the prompt for each task type.

\begin{table}[h]
\centering
\caption{Task-specific prompt instructions and answer formats.}
\renewcommand{\arraystretch}{1.2}
\label{tab:task_instructions}
\resizebox{\columnwidth}{!}{%
\begin{tabular}{p{4.5cm}p{5cm}c}
\toprule
\textbf{Task Type} & \textbf{Key Instruction} & \textbf{Answer} \\
\midrule
\texttt{pairwise\_configuration} & Focus on relative positions, orientations, left/right, front/behind & Yes/No \\
\texttt{object\_rel\_direction} & Analyze relative positions from object-centric perspective & A/B/C/D \\
\texttt{object\_rel\_distance} & Compare closest-point distances to multiple candidates & A/B/C/D \\
\texttt{object\_abs\_distance} & Calculate min distance between closest bbox points & Numeric \\
\texttt{object\_counting} & Count objects matching specified category & Numeric \\
\texttt{object\_size\_estimation} & Report max(x\_size, y\_size, z\_size) in requested unit & Numeric \\
\texttt{room\_size\_estimation} & Estimate floor area from wall bounding boxes & Numeric \\
\texttt{obj\_appearance\_order} & Determine temporal order of first appearance in video & A/B/C/D \\
\texttt{route\_planning} & Plan navigation with turn directions at waypoints & A/B/C/D \\
\bottomrule
\end{tabular}%
}
\end{table}

\paragraph{Example Prompt.}
Figure~\ref{fig:prompt_example} shows a complete prompt for the \texttt{object\_rel\_direction} task.

\begin{figure}[h]
\centering
\begin{tcolorbox}[
    colback=gray!5,
    colframe=gray!60,
    boxrule=0.5pt,
    arc=2pt,
    left=3pt, right=3pt, top=3pt, bottom=3pt,
    width=\columnwidth
]
\scriptsize
\texttt{You are a multimodal reasoning model that interprets structured scene inputs. You will be provided with a list of bounding boxes representing objects in the scene...}\\[3pt]
\texttt{\textbf{Coordinate System Conventions:}}\\
\texttt{1. World Frame: Z-axis points Up. X and Y are horizontal.}\\
\texttt{2. Yaw: 0 rad is along +X. Positive Yaw is counter-clockwise.}\\[3pt]
\texttt{\textbf{Measurement Definitions:}}\\
\texttt{1. Object Size: max(x\_size, y\_size, z\_size)}\\
\texttt{2. Absolute Distance: minimum Euclidean distance between closest points}\\[3pt]
\texttt{\textbf{IMPORTANT - Reading bbox\_3d values:}}\\
\texttt{The bbox\_3d array format is: [x\_center, y\_center, z\_center, x\_size, y\_size, z\_size, roll, pitch, yaw]}\\
\texttt{- The YAW is the 9th (last) value - check it carefully!}\\[3pt]
\hrule
\vspace{3pt}
\texttt{The bounding boxes are:}\\
\texttt{\{"label":"bathtub",\\ "bbox\_3d":[-1.284,1.547,0.267,0.812,1.701,0.534,0,0,0]\}}\\
\texttt{\{"label":"cabinet",\\ "bbox\_3d":[-0.065,2.001,1.003,0.599,0.593,1.857,0,0,0]\}}\\
\texttt{\{"label":"washer",\\ "bbox\_3d":[0.632,2.103,0.429,0.609,0.571,0.858,0,0,0]\}}\\[3pt]
\hrule
\vspace{3pt}
\texttt{\textbf{Task Type: Object Relative Direction}}\\
\texttt{You are analyzing the Object Relative Direction between two objects. Focus on their relative positions, orientations...}\\[3pt]
\hrule
\vspace{3pt}
\texttt{Based on the provided bounding boxes, answer the following question:}\\
\texttt{\textbf{Standing at the bathtub, facing the cabinet, is the washer to your front-left, front-right, back-left, or back-right?}}\\[2pt]
\texttt{Options: A. front-right, B. back-right, C. front-left, D. back-left}\\[3pt]
\texttt{Think step by step. Keep your reasoning concise (under 1000 words).}\\
\texttt{IMPORTANT: End your response with exactly:}\\
\texttt{Final Answer: A ~or~ Final Answer: B ~or~ Final Answer: C ~or~ Final Answer: D}
\end{tcolorbox}
\caption{Complete prompt example for the \texttt{object\_rel\_direction} task.}
\label{fig:prompt_example}
\end{figure}

\subsection{Reward Function Details}
\label{app:rubric}

Our reward function (Eq.~\ref{eq:reward}) consists of three components: accuracy, format compliance, and spatial rubrics. Table~\ref{tab:reward_summary} provides an overview of each component and its range.

\begin{table}[h]
\centering
\caption{Reward function components and their ranges.}
\renewcommand{\arraystretch}{1.2}
\label{tab:reward_summary}
\resizebox{\columnwidth}{!}{%
\begin{tabular}{lcl}
\toprule
\textbf{Component} & \textbf{Range} & \textbf{Description} \\
\midrule
$r_{\text{acc}}$ & $\{0, 1\}$ & Exact match with ground truth $a^*$ \\
$r_{\text{format}}$ & $[-0.7, +0.1]$ & Structure compliance and penalties \\
$r_{\text{rubric}}$ & $[-0.65, +0.8]$ & Task-specific reasoning quality \\
\midrule
$r_{\text{total}}$ & $[-0.5, 1.8]$ & Clipped sum of all components \\
\bottomrule
\end{tabular}%
}
\end{table}

\paragraph{Format Compliance ($r_{\text{format}}$).}
Table~\ref{tab:format_rubrics} details the format compliance indicators that reward proper response structure and penalize degenerate outputs.

\begin{table}[h]
\centering
\renewcommand{\arraystretch}{1.2}
\caption{Format compliance indicators.}
\label{tab:format_rubrics}
\resizebox{0.85\columnwidth}{!}{%
\begin{tabular}{p{4cm}p{4cm}c}
\toprule
\textbf{Indicator} & \textbf{Condition} & \textbf{Weight} \\
\midrule
Valid format & Single ``Final Answer:'' with valid response & $+0.1$ \\
Missing format & No ``Final Answer:'' detected & $-0.2$ \\
Degenerate repetition & Pathological output patterns (e.g., repeated phrases) & $-0.5$ \\
Excessive length & Response $> 4000$ characters & $\leq -0.3$ \\
Too short & Response $< 200$ characters & $-0.2$ \\
\bottomrule
\end{tabular}%
}
\end{table}

\paragraph{Base Reasoning Rubrics.}
Table~\ref{tab:base_rubrics} lists reasoning quality indicators applied across all spatial reasoning tasks.

\begin{table}[h]
\centering
\caption{Base reasoning quality indicators (all tasks).}
\label{tab:base_rubrics}
\renewcommand{\arraystretch}{1.2}
\resizebox{\columnwidth}{!}{%
\begin{tabular}{p{3.5cm}p{5cm}c}
\toprule
\textbf{Indicator $\phi_i$} & \textbf{Condition} & \textbf{$w_i$} \\
\midrule
Structured reasoning & $\geq 2$ step indicators (``Step 1'', ``First'', ``Then'') & $+0.1$ \\
Conclusion statement & Contains conclusion phrase (``therefore'', ``thus'') & $+0.1$ \\
Reasoning consistency & Analysis logically supports final answer & $+0.1$ \\
Reasoning inconsistency & Analysis contradicts final answer & $-0.3$ \\
\bottomrule
\end{tabular}%
}
\end{table}

\paragraph{Task-Specific Rubrics.}
We define specialized indicators for each task type. Tables~\ref{tab:direction_rubrics} and \ref{tab:config_rubrics} show rubrics for two representative tasks: \texttt{object\_rel\_direction} and \texttt{pairwise\_configuration}.

\begin{table}[h]
\centering
\caption{Task-specific rubrics for \texttt{object\_rel\_direction}.}
\label{tab:direction_rubrics}
\renewcommand{\arraystretch}{1.2}
\resizebox{\columnwidth}{!}{%
\begin{tabular}{p{4cm}p{4cm}c}
\toprule
\textbf{Indicator $\phi_i$} & \textbf{Condition} & \textbf{$w_i$} \\
\midrule
\multicolumn{3}{l}{\textit{Positive Indicators}} \\
Coordinate extraction & Extracts $\geq 3$ coordinate values & $+0.10$ \\
Facing vector computation & Computes observer-to-target direction & $+0.15$ \\
Local coordinate system & Constructs forward/right basis vectors & $+0.25$ \\
Vector projection & Uses dot product for local components & $+0.20$ \\
Quadrant determination & Maps component signs to quadrant & $+0.10$ \\
\midrule
\multicolumn{3}{l}{\textit{Negative Indicators}} \\
World-coordinate confusion & Uses $+x \to \text{right}$ without transform & $-0.25$ \\
Missing transformation & Skips local frame construction & $-0.20$ \\
Incorrect rotation & Wrong formula for right vector & $-0.10$ \\
Incomplete 2D analysis & Checks only one axis & $-0.10$ \\
Lucky guess & Correct answer without proper method & $-0.30$ \\
\bottomrule
\end{tabular}%
}
\end{table}

\begin{table}[h]
\centering
\caption{Task-specific rubrics for \texttt{pairwise configuration}.}
\label{tab:config_rubrics}
\renewcommand{\arraystretch}{1.2}
\resizebox{\columnwidth}{!}{%
\begin{tabular}{p{4cm}p{4cm}c}
\toprule
\textbf{Indicator $\phi_i$} & \textbf{Condition} & \textbf{$w_i$} \\
\midrule
\multicolumn{3}{l}{\textit{Positive Indicators}} \\
Perspective-based reasoning & Uses object's perspective, not viewer's & $+0.15$ \\
Orientation awareness & Considers yaw angle in analysis & $+0.15$ \\
Coordinate analysis & Compares object coordinates correctly & $+0.10$ \\
Directional consistency & Handles opposite pairs (left/right) & $+0.10$ \\
\midrule
\multicolumn{3}{l}{\textit{Negative Indicators}} \\
Viewer-centric error & Uses viewer perspective instead of object & $-0.15$ \\
Orientation ignorance & Ignores yaw when determining directions & $-0.15$ \\
\bottomrule
\end{tabular}%
}
\end{table}

\subsection{Rubric Scoring Example}
\label{app:reward_example}

Figure~\ref{fig:rubric_example} illustrates how the rubric reward distinguishes between correct answers achieved through proper spatial reasoning versus coincidental correctness. Both responses arrive at the same correct answer, but receive significantly different rewards based on their reasoning methodology.

\begin{figure*}[ht!]
\centering
\begin{tcolorbox}[
    colback=white,
    colframe=black!70,
    boxrule=0.6pt,
    arc=2pt,
    width=\textwidth
]
\small
\begin{minipage}[t]{0.32\textwidth}
\textbf{\textcolor{green!50!black}{$\checkmark$~Proper Reasoning}}\\[3pt]
\scriptsize
\textit{\textbf{Step 1: Extract coordinates.}}\\
\textit{Bathtub: $(-1.28, 1.55)$}\\
\textit{Cabinet: $(-0.07, 2.00)$}\\
\textit{Washer: $(0.63, 2.10)$}\\[2pt]
\textit{\textbf{Step 2: Compute facing direction.}}\\
\textit{$\vec{v} = (-0.07, 2.00) - (-1.28, 1.55)$}\\
\textit{$= (1.21, 0.45)$}\\
\textit{Normalize: $\hat{f} = (0.94, 0.35)$}\\[2pt]
\textit{\textbf{Step 3: Build local coordinates.}}\\
\textit{forward $= \hat{f} = (0.94, 0.35)$}\\
\textit{right $= (\hat{f}_y, -\hat{f}_x) = (0.35, -0.94)$}\\[2pt]
\textit{\textbf{Step 4: Project washer position.}}\\
\textit{$\vec{w} = (0.63, 2.10) - (-1.28, 1.55)$}\\
\textit{$= (1.91, 0.55)$}\\
\textit{front $= \vec{w} \cdot \hat{f} = 1.99 > 0$ {\color{green!50!black}$\checkmark$}}\\
\textit{right $= \vec{w} \cdot \hat{r} = 0.15 > 0$ {\color{green!50!black}$\checkmark$}}\\[2pt]
\textit{Therefore, washer is \textbf{front-right}.}\\
\textit{Final Answer: A}
\end{minipage}%
\hfill
\begin{minipage}[t]{0.32\textwidth}
\textbf{\textcolor{red!60!black}{$\times$~Flawed Reasoning}}\\[3pt]
\scriptsize
\textit{Looking at the coordinates:}\\
\textit{Washer is at $(0.63, 2.10)$}\\[2pt]
\textit{In world coordinates:}\\
\textit{$x = 0.63 > 0 \Rightarrow$ right}\\
\textit{$y = 2.10 > 0 \Rightarrow$ front}\\[2pt]
\textit{So washer is front-right.}\\
\textit{Final Answer: A}\\[8pt]
\textcolor{red!60!black}{\textbf{Error:} Uses world coordinate}\\
\textcolor{red!60!black}{axes directly ($+x \to$ right)}\\
\textcolor{red!60!black}{without constructing local}\\
\textcolor{red!60!black}{reference frame from observer's}\\
\textcolor{red!60!black}{position and facing direction.}\\[4pt]
\textcolor{red!60!black}{This approach fails when the}\\
\textcolor{red!60!black}{observer faces a different direction.}
\end{minipage}%
\hfill
\begin{minipage}[t]{0.35\textwidth}
\centering
\textbf{Reward Breakdown}\\[3pt]
\scriptsize
\renewcommand{\arraystretch}{1.2}
\begin{tabular}{@{}l|rr@{}}
\textbf{Component} & \textbf{P} & \textbf{F} \\
\hline
$r_{\text{acc}}$ (correct) & $+1.0$ & $+1.0$ \\
$r_{\text{fmt}}$ (format) & $+0.1$ & $+0.1$ \\
Structured reason. & $+0.1$ & $0$ \\
Coord. extraction & $+0.1$ & $+0.05$ \\
Facing vector & $+0.15$ & $0$ \\
Local coord. sys. & $+0.25$ & $0$ \\
Vector projection & $+0.2$ & $0$ \\
World-coord. conf. & $0$ & $-0.25$ \\
Lucky guess pen. & $0$ & $-0.3$ \\
\hline
\textbf{Total} & $\mathbf{+1.8}$ & $\mathbf{+0.6}$ \\
\end{tabular}
\end{minipage}
\end{tcolorbox}
\caption{Rubric scoring comparison for \texttt{object\_rel\_direction}. Both responses achieve the correct answer (A: front-right), but the rubric reward assigns $3\times$ higher reward to proper spatial reasoning methodology.}
\label{fig:rubric_example}
\end{figure*}

The ``lucky guess'' penalty ($w=-0.30$) is applied when the model arrives at the correct answer but fails to demonstrate proper spatial reasoning methodology---specifically, when $r_{\text{acc}}=1$ but neither the local coordinate system indicator nor the vector projection indicator is satisfied. This design choice encourages learning of transferable reasoning skills that generalize across different spatial configurations.

\newpage
\clearpage

\section{Parameter Count Clarification and Scaling Analysis}
\label{sec:appendix_param}

\subsection{Model Parameter Breakdown}

Our full model introduces an additional spatial encoder $\phi$ beyond the base VLM. The complete parameter breakdown is as follows:

\begin{table}[ht!]
\centering
\renewcommand{\arraystretch}{1.2}
\caption{Parameter breakdown of our full model.}
\resizebox{0.7\linewidth}{!}{
\begin{tabular}{lc}
\toprule
\textbf{Component} & \textbf{Parameters} \\
\midrule
SAM2 Encoder (Hiera-ViT-L) & 214M \\
DA3 Encoder (ViT-G) & 1.43B \\
Qwen3-VL-4B (LM backbone) & $\sim$4B \\
\midrule
\textbf{Total} & $\sim$\textbf{5.6B} \\
\bottomrule
\end{tabular}
}
\label{tab:param_breakdown}
\vspace{-5mm}
\end{table}

A natural question arises: \textit{given a fixed parameter budget, is it more effective to scale the spatial encoder or the language model?} We investigate this through the following experiments.

\subsection{Parameter Allocation: Spatial Encoder vs. Language Model}

\noindent\textbf{Setup.} We compare our model against Qwen3-VL-8B, which allocates an additional 4B parameters to the language model rather than the spatial encoder. This comparison isolates the effect of parameter allocation strategy. $\dagger$ denotes that we use the subsets for analysis (\texttt{Object\_rel\_direction} for VSI-Bench and \texttt{Pairwise\_configuration} for Video-RoboSpatial).

\begin{table}[h]
\centering
\renewcommand{\arraystretch}{1.2}
\vspace{-2mm}
\caption{Parameter allocation comparison. Despite fewer total parameters ($\sim$5.6B vs. $\sim$8B), allocating capacity to the spatial encoder outperforms scaling the language model.}
\resizebox{\linewidth}{!}{
\begin{tabular}{lcccc}
\toprule
\textbf{Model} & \textbf{Total Param.} & \textbf{Spatial Enc.} & \textbf{LM} & \textbf{VSI-Bench} (Avg 8)   \\
\midrule
Qwen3-VL-8B & 8B &  \textbf{0} & 8B & 55.1  \\
Ours & \textbf{5.6B} & 1.6B &  \textbf{4B} & \textbf{60.8} (+2.9\%) \\
\bottomrule
\end{tabular}
}
\label{tab:param_allocation}
\end{table}

\begin{tcolorbox}[colback=gray!10, colframe=gray!50, title=Finding 1]
\textit{Allocating parameters to the spatial encoder yields greater improvements on spatial reasoning tasks than scaling the language model.} Despite having fewer total parameters, our model outperforms Qwen3-VL-8B by 2.9\% average accuracy on VSI-bench suggesting that spatial understanding is bottlenecked by encoder capacity rather than language modeling capacity.
\end{tcolorbox}

\section{Spatial Encoding Details}
\label{sec:app-det-details}
In this section, we provide additional implementation details about multi-frame fusion strategy (\S\ref{sec:app-det-fusion}), 3D head design (\S\ref{sec:app-det-head}) and training objectives (\S\ref{sec:app-det-loss}).

\subsection{Multi-Frame 3D Detection Fusion}
\label{sec:app-det-fusion}
Our method predicts 3D bounding boxes in the camera coordinate system for each RGB frame. To obtain scene-level predictions, we transform per-frame detections into a unified world coordinate system and merge overlapping predictions across frames. This section details the transformation pipeline and spatial clustering strategy.

Given a sequence of $T$ RGB frames with known camera poses $\{\mathbf{T}_1, \mathbf{T}_2, \ldots, \mathbf{T}_T\}$, our Spatial Encoder predicts spatial codes for each frame in its respective camera coordinate system. For frame $t$, the output is $\mathbf{c}^t = \{\mathbf{c}_i^t\}_{i=1}^{n_t}$, where each spatial code is:
\begin{equation}
    \mathbf{c}_i^t = (l_i^t, \mathbf{p}_i^t, \mathbf{s}_i^t, \mathbf{r}_i^t),
\end{equation}
with semantic label $l_i^t$, position $\mathbf{p}_i^t \in \mathbb{R}^3$, size $\mathbf{s}_i^t \in \mathbb{R}^3$, and orientation (quaternion) $\mathbf{r}_i^t \in \mathbb{R}^4$.

The goal is to aggregate these per-frame predictions into a single set of scene-level spatial codes $\mathbf{c}^* = \{\mathbf{c}_i^*\}_{i=1}^{n^*}$ in world coordinates, where $n^* \ll \sum_{t=1}^T n_t$.

\paragraph{Spatial Clustering for Code Fusion}
Two spatial codes $\mathbf{c}_i$ and $\mathbf{c}_j$ (in world coordinates) are considered to represent the same object if they satisfy both:
\begin{itemize}
    \item{Spatial Proximity.}
    The Euclidean distance between their positions is below a threshold:
    \begin{equation}
        \|\mathbf{p}_i - \mathbf{p}_j\| < \tau_{\text{dist}}
    \end{equation}
    where $\tau_{\text{dist}} = 0.3$ meters. This criterion quickly filters out clearly distinct objects.
    \item{Geometric Overlap.}
    Among spatially proximate pairs, we require sufficient 3D overlap:
    \begin{equation}
        \text{IoU}_{3D}(\mathbf{c}_i, \mathbf{c}_j) > \tau_{\text{iou}},
    \end{equation}
    where $\tau_{\text{iou}} = 0.3$. This ensures that only geometrically consistent detections are merged.

\end{itemize}

\paragraph{3D IoU Computation}

For oriented 3D bounding boxes defined by position $\mathbf{p}$, size $\mathbf{s} = [w, h, l]^T$, and orientation $\mathbf{r}$, we compute the Intersection over Union as:
\begin{equation}
    \text{IoU}_{3D} = \frac{V_{\text{inter}}}{V_1 + V_2 - V_{\text{inter}}}.
\end{equation}

The intersection volume $V_{\text{inter}}$ is computed by:
\begin{enumerate}
    \item Projecting both boxes onto the ground plane (bird's eye view) to compute the 2D intersection area $A_{\text{BEV}}$ using convex hull intersection
    \item Computing the vertical overlap height $h_{\text{overlap}}$ between the two boxes along the $y$-axis
    \item Computing intersection: $V_{\text{inter}} = A_{\text{BEV}} \times h_{\text{overlap}}$
\end{enumerate}

Additionally, individual box volumes are computed from their sizes: $V = w \times h \times l$.

\paragraph{Cluster Aggregation}
For each cluster of $N$ detections $\{\mathbf{c}_1, \ldots, \mathbf{c}_N\}$ identified as representing the same object, we compute the representative spatial code by averaging their parameters at the scene-level:
\begin{equation}
    \begin{aligned}
    \bar{\mathbf{p}} &= \frac{1}{N}\sum_{i=1}^N \mathbf{p}_i, \\
    \bar{\mathbf{s}} &= \frac{1}{N}\sum_{i=1}^N \mathbf{s}_i, \\
    \bar{\mathbf{r}} &= \text{QuaternionMean}(\{\mathbf{r}_1, \ldots, \mathbf{r}_N\}),
\end{aligned}
\end{equation}
where orientations (quaternions) are averaged with proper handling of quaternion sign ambiguity. The semantic label $\bar{l}$ is selected by majority voting among $\{l_1, \ldots, l_N\}$.

\subsection{Evaluation Protocol}

For evaluation against ground truth annotations, we perform bipartite matching between predicted and ground truth spatial codes using the Hungarian algorithm. The matching is based on maximizing the total 3D IoU:
\begin{equation}
    \text{maximize} \sum_{i,j} \text{IoU}_{3D}(\mathbf{c}_i^{\text{pred}}, \mathbf{c}_j^{\text{gt}}) \cdot m_{ij},
\end{equation}
where $m_{ij} \in \{0, 1\}$ indicates whether prediction $i$ is matched to ground truth $j$.

A matched pair is considered a true positive if:
\begin{equation}
    \text{IoU}_{3D}(\mathbf{c}_i^{\text{pred}}, \mathbf{c}_j^{\text{gt}}) \geq \tau_{\text{eval}},
\end{equation}
where $\tau_{\text{eval}} \in \{0.25, 0.50\}$ depending on the evaluation metric (F1@25 or F2@50).

\subsection{3D Head Design}
\label{sec:app-det-head}
Our 3D Head composes a series of transformer layers and MLPs predicting multiple spatial attributes through specialized prediction heads. 

\begin{figure}[h]
    \centering
    \includegraphics[width=\linewidth]{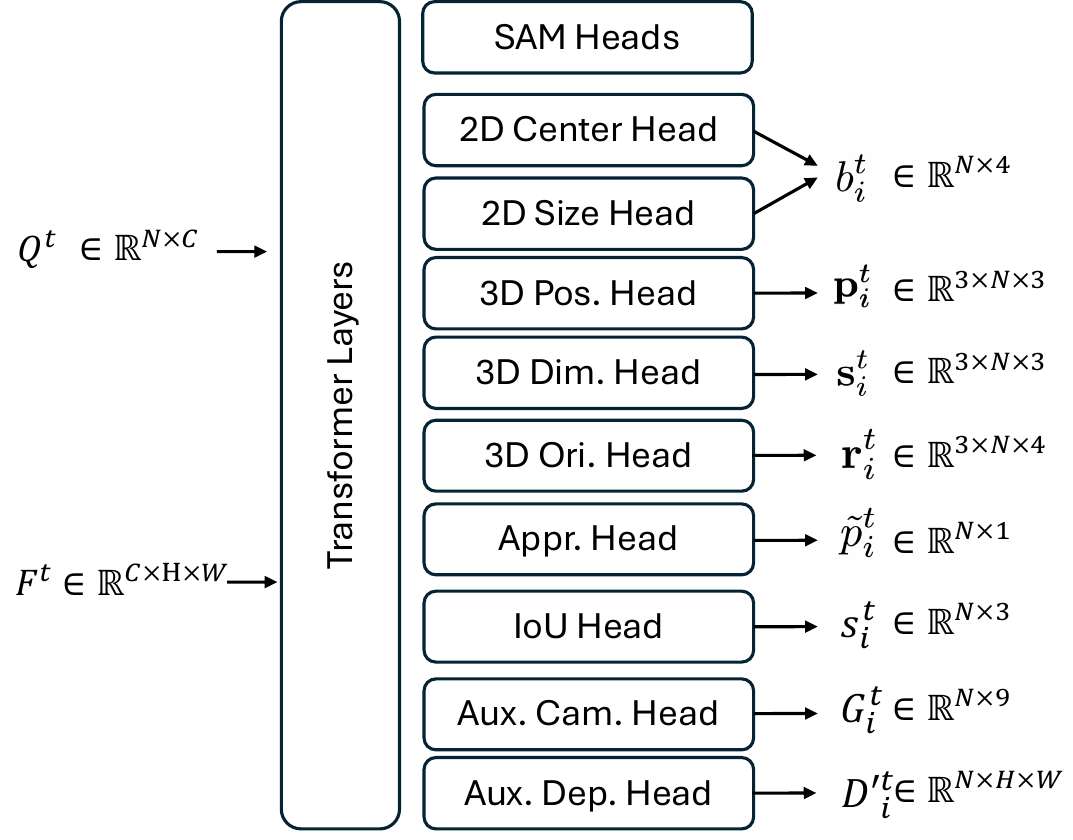}
    \caption{The details of 3D Head implementation.}
    \label{fig:app-det-heads}
\end{figure}
We retain SAM's original mask heads and use it for tracking, as implemented in~\cite{ravi2024sam}. The detection heads predict 3D bounding boxes: a 2D box head predicts normalized center coordinates and log-space width/height for 2D bounding box $b^t_i$; per-mask 3D box heads output scaled $(x_i^t,y_i^t)$ translation, log-space Z depth $(z^t_i)$ and dimensions $(sx_i^t,sy_i^t,sz_i^t)$, and confidence $c_i^t$; orientation is predicted through normalized quaternion representations. All spatial predictions ($x_i^t,y_i^t,z^t_i,sx_i^t,sy_i^t,sz_i^t$) are scaled by a learned scale factor to handle metric ambiguity. Following SAM~\cite{kirillov2023sam,ravi2024sam}, we implement three separate heads with an additional IoU head to output the confidence of each head. During inference, we choose the prediction with highest confidence.

We also include an auxiliary camera head and auxiliary depth head for geometry supervision, enforcing the feature to be aware of spatial information conditioned on $h_\phi^{dep}(F_t)$. This auxiliary camera prediction and depth map prediction are fused with the output of depth head by a factor of 0.1.

\subsection{Training Objective Details}
\label{sec:app-det-loss}
The detection loss $\mathcal{L}_{\text{Detection}}$ supervises frame-wise bounding box prediction. 
Following~\cite{zhang2025detect,brazil2023omni3d,yang20253d}, the 2D detection loss is a combination of GIoU loss and L1 loss between predicted 2D boxes and ground-truth 2D boxes, as:
\begin{equation}
    \mathcal{L}_{\text{2D det}}=\lambda_1 \cdot (1-GIoU(\hat b^t_{2D}, b^t_{2D}))+\lambda_2 \cdot \|\hat b^t_{2D}- b^t_{2D}\|.
\end{equation}
For position $\mathbf{p}^t_i=(x_i^t,y_i^t,z_i^t)$, we project $(x_i^t,y_i^t)$ in to pixel space by predicted intrinsics $K$, and use L1 loss to align it with ground-truth object center $(\hat x_i,\hat y_i)$ in pixel space, while aligning predicted $z_i^t$ with the real value $\hat z_i$ by a Laplacian Aleatoric Uncertainty loss: 
\begin{equation}
\begin{aligned}
    \mathcal{L}_{\text{pos}}=&\lambda_3\cdot\|(\hat x_i^t,\hat y_i^t)-K\cdot(x_i^t,y_i^t)\| \\
    &+\lambda_4\cdot(\sqrt{2} \cdot e^{-u_z} \cdot \|\hat z_i - z_i^t\| + u_z).
\end{aligned}
\end{equation}
For object size prediction, L1 losses are applied to between predicted dimensions and ground-truth dimensions. For orientation, the prediction is represented as quaternions and L1 loss is applied after normalization. To ensure corner-wise alignment, a chamfer loss is implemented:
\begin{equation}
\mathcal{L}_{\text{chamfer}} = \frac{1}{|\mathcal{C}_i|}\sum_{\mathbf{c} \in \mathcal{C}_i} \min_{\hat{\mathbf{c}} \in \hat{\mathcal{C}}_i} \|\mathbf{c} - \hat{\mathbf{c}}\|_2 + \frac{1}{|\hat{\mathcal{C}}_i|}\sum_{\hat{\mathbf{c}} \in \hat{\mathcal{C}}_i} \min_{\mathbf{c} \in \mathcal{C}_i} \|\hat{\mathbf{c}} - \mathbf{c}\|_2,
\end{equation}
where $\mathcal{C}_i$ and $\hat{\mathcal{C}}_i$ denote the sets of predicted and ground-truth 3D bounding box corners for object $i$, respectively. To select the most confident heaf out of three predicted heads, we train an addtional head to predict the 3D IoU score of each head:
\begin{equation}
    \mathcal{L}_{\text{3D-IoU}} = \frac{1}{|\mathcal{V}|} \sum_{i \in \mathcal{V}} \sum_{k=1}^{3} \left\| \tilde{s}_i^{k} - \text{IoU}_{\text{3D}}(\mathcal{B}_{i,k}^{t}, \hat{\mathcal{B}}_i^{t}) \right\|_1,
\end{equation}
where each $\mathcal{B}_i^t$ is calculated from $(\mathbf{p}^t_i, \mathbf{s}^t_i, \mathbf{r}^t_i)$.

The geometry loss supervises depth prediction and camera pose prediction. Following~\cite{wang2025vggt,lin2025depth,wang2024dust3r}, we use scale-invariant depth loss with an aleatoric-uncertainty term and a gradient term for depth map prediction:
\begin{equation}
\begin{aligned}
    \mathcal{L}_{\text{depth}} =& \sqrt{\frac{1}{H\times W}\sum_{i} \left(e^{D^t_{\text{conf},i}} g_i^2 - D^t_{\text{conf},i}\right) - \lambda \bar{g}^2 + \alpha} \\
    &+ \|\tilde{D}^t- \hat{\tilde{D}}^t\|,
\end{aligned}
\end{equation}
where $g_i = \log(\tilde{D}^t_i + \alpha) - \log(\hat{\tilde{D}}^t_i + \alpha)$ denotes the log-depth difference at pixel $i$, $\bar{g} = \frac{1}{N}\sum_{i} g_i$ is its mean value, and $\tilde{D}^t = D^t/(\|D^t\|_2+1)$ represents the normalized predicted depth map. Here $D^t_{\text{conf}}$ is the predicted confidence map that models aleatoric uncertainty, $\lambda$ controls the scale-invariance penalty, and $\alpha=10^{-7}$ ensures numerical stability.

The camera pose is encoded as $\mathbf{e}^t = [\mathbf{t}^t, \mathbf{q}^t, f^t]$, where $\mathbf{t}^t \in \mathbb{R}^3$ represents translation, $\mathbf{q}^t \in \mathbb{R}^4$ is the rotation quaternion, and $f^t$ denotes the focal length. The camera loss consists of three components:
\begin{equation}
    \mathcal{L}_{\text{camera}} 
    =\lambda_T\cdot \|\hat{\mathbf{t}}^t - \mathbf{t}^t\|_1 + \lambda_R \cdot \|\hat{\mathbf{q}}^t - \mathbf{q}^t\|_1 + \lambda_f\cdot \|\hat{f}^t - f^t\|_1,
\end{equation}
with $\lambda_T$, $\lambda_R$, and $\lambda_f$ being the loss weights for translation, rotation, and focal length respectively. Here $\hat{\cdot}$ denotes ground truth values.

The tracking loss $\mathcal{L}_{\text{tracking}}$ supervises whether each object appears in the current frame, implemented as a sigmoid focal loss:
\begin{equation}
    \mathcal{L}_{\text{tracking}} = -\frac{1}{N} \sum_{i=1}^{N} \alpha_t (1 - p_t)^\gamma \cdot \text{BCE}(\tilde{p}_i^t, y_i),
\end{equation}
where $\tilde{p}_i^t$ is the predicted appearance logit and $y_i \in \{0,1\}$ indicates object presence.

\subsection{Training-Free Implementation}
\label{sec:app-encoder-v2}

\end{document}